\documentclass[10pt,twocolumn,letterpaper]{article}

\usepackage{iccv}
\usepackage{times}
\usepackage{epsfig}
\usepackage{graphicx}
\usepackage{amsmath}
\usepackage{amssymb}

\usepackage[pagebackref=true,breaklinks=true,letterpaper=true,colorlinks,bookmarks=false]{hyperref}

\usepackage{etoolbox}
\usepackage{bm}
\usepackage{amsthm}%liwen
\usepackage{algorithm}%liwen
\usepackage{algorithmic}%liwen
\usepackage{subfigure}%liwen

\newcommand{\MyMapTemplatePrefixc}[4]{\expandafter#1\csname#3#4\endcsname{#2{#4}}} % it remembles a template: \#3#4 --> #2{#4}
\forcsvlist{\MyMapTemplatePrefixc {\def} {\mathcal}{c}} {A,B,C,D,E,F,G,H,I,J,K,L,M,N,O,P,Q,R,S,T,U,V,W,X,Y,Z}  % e.g., \cA --> \mathcal{A}

\newcommand{\MyMapTemplatePrefixtb}[5]{\expandafter#1\csname#4#5\endcsname{#2{#3{#5}}}} % it remembles a template: \#3#4 --> #2{#4}
\forcsvlist{\MyMapTemplatePrefixtb {\def} {\tilde}{\mathbf}{t}} {A,B,C,D,E,F,G,H,I,J,K,L,M,N,O,P,Q,R,S,T,U,V,W,X,Y,Z}  % e.g., \tA --> \tilde{A}
\forcsvlist{\MyMapTemplatePrefixtb {\def} {\tilde}{\mathbf}{t}} {0,1,a,b,c,d,e,f,g,h,i,j,k,l,m,n,o,p,q,r,s,u,v,w,x,y,z}  % e.g., \ta --> \tilde{a}

\newcommand{\MyMapTemplateNoPrefix}[3]{\expandafter#1\csname#3\endcsname{#2{#3}}}
\forcsvlist{\MyMapTemplateNoPrefix {\def} {\mathbf} } {0,1,a,b,c,d,e, f, g, h, i, j, k, l, m, n, o, p, q, r, u, v, w, x, y, z} % e.g., \a --> \mathbf{a}
\forcsvlist{\MyMapTemplateNoPrefix {\def} {\mathbf} } {A,B,C,D,E,F,G,H,I,J,K,L,M,N,O,P,Q,R,S,T,U,V,W,X,Y,Z}  % e.g., \A --> \mathcal{A}

\def\etal{\emph{et al.}\@\xspace}
\def\resp{\emph{resp.}\@\xspace}

\usepackage{booktabs} % Nice tables
\usepackage[table,dvipsnames]{xcolor}
\definecolor{rowblue}{RGB}{220,230,240}

\iccvfinalcopy
\ificcvfinal\fi

\begin{document}

%%%%%%%%% TITL
\title{WebVision Database: Visual Learning and Understanding from Web Data}

\author{Wen Li$^1$, Limin Wang$^1$, Wei Li$^2$, Eirikur Agustsson$^1$, Luc Van Gool$^{1}$\\
$^1$ Computer Vision Laboratory, ETH Zurich\\
$^2$ Google Switzerland\\
}

\maketitle

%%%%%%%%% ABSTRACT
\begin{abstract}
In this paper, we present a study on learning visual recognition models from large scale noisy web data. We build a new database called \emph{WebVision}, which contains more than $2.4$ million web images crawled from the Internet by using queries generated from the $1,000$ semantic concepts of the ILSVRC 2012 benchmark. Meta information along with those web images (\eg, title, description, tags, \etc) are also crawled. A validation set and test set containing human annotated images are also provided to facilitate algorithmic development. Based on our new database, we obtain a few interesting observations: 1) the noisy web images are sufficient for training a good deep CNN model for visual recognition; 2) the model learnt from our WebVision database exhibits comparable or even better generalization ability than the one trained from the ILSVRC 2012 dataset when being transferred to new datasets and tasks; 3) a domain adaptation issue (a.k.a., dataset bias) is observed, which means the dataset can be used as the largest benchmark dataset for visual domain adaptation. Our new WebVision database and relevant studies in this work would benefit the advance of learning state-of-the-art visual models with minimum supervision based on web data.
\end{abstract}

%%%%%%%%% BODY TEXT
\section{Introduction}
The recent success of deep learning has shown that a deep architecture in conjunction with abundant quantities of labeled training data is the most promising approach for most vision tasks~\cite{krizhevsky2012imagenet,HeZRS16,SzegedyLJSRAEVR15,long2015fully,ren2015faster, DosovitskiyFIHH15,TranBFTP15,SimonyanZ14,WangXW0LTG16,DongLHT16}. However, annotating a large-scale dataset for training such deep neural networks is costly and time-consuming, even with the availability of scalable crowd-sourcing platforms like Amazon Mechanical Turk. As a result, there are relatively few public large-scale datasets (\eg, ImageNet~\cite{ImageNet} and Places2~\cite{Places2}) from which it is possible to learn generic visual representations from scratch.

Thus, it is unsurprising that there is a continued interest in developing novel deep learning systems trained on low-cost data, including unlabeled images/videos~\cite{Vincent10, Radford15}, self-supervised and semi-supervised approaches~\cite{Doersch15, Wang15, Agrawal15, Noroozi16}, and methods that exploit weak and noisy labels from auxiliary sources~\cite{Chen15, Joulin16, Owens16, Krause16}. In particular, there is promising recent work on using the web as a source of supervision for learning deep representations for a variety of important computer vision applications, including image annotation, object detection, and fine-grained classification~\cite{ Chen15, Joulin16, Krause16}.

Learning from web data differs from purely supervised or unsupervised learning because images and videos on the web are naturally accompanied with abundant meta data (such as surrounding text, title, tags, \etc) that can provide weak supervision without the tedium or expense of crowd-sourced manual label. While the existing works~\cite{Vijayanarasimhan08,  Li14, Chen15, Joulin16, Krause16} have shown advantages of using web data in various applications, their tasks and methodologies differ from each other, making it hard to identify key issues and effective ways when utilizing web data. Moreover, their results were often obtained using much more images or categories, making it difficult to understand the capacity of noisy web images for learning visual recognition models when compared with the human-annotated datasets.

To address these problems, we present a rigorous study  on learning visual recognition models from large scale noisy web data.  We build a new web image database called {\emph{WebVision}},  which contains more than $2.4$ million of web images crawled from the Internet (about $1$ million from Google Image search, and $1.4$ million from Flickr)  by using queries generated from the same $1,000$ semantic concepts as the benchmark ILSVRC 2012 datast. Meta information along with those web images (\eg, title, description, tags, \etc) are also crawled. A validation set and a test set, each containing $50,000$ human annotated images, are also provided to facilitate algorithmic development. With this new database, we are keen to answer the following questions:
\begin{itemize}
\item \emph{How do the noisy labels of web images impact the visual recognition models, compared with those from human-annotated data?} The WebVision database is constructed using the same $1,000$ semantic concepts as the ILSVRC 2012 datast, which is the current most popular large scale human-annotated benchmark dataset for image classification. This allows us to compare directly with the ILSVRC 2012 dataset on the $1,000$ categories image classification task. Our experimental results show that the CNN model learnt from WebVision achieves quite competitive results with the one learnt from ILSVRC 2012. Further analysis, by varying the number of training images, indicates that the web images are limited in the quality of labels (\ie, the label noise in the web data), but can be compensated by the advantage in the large quantity.
\item \emph{How good is the generalization ability of the learnt models from web images, when applied to other datasets and tasks?} When applying the model trained on WebVision (\resp, ILSVRC 2012) to the ILSVRC 2012 (\resp, WebVision) validation set, we observe a performance drop when compared with when applying the model to the WebVision (\resp, ILSVRC 2012) validation set. This indicates that there exists a dataset bias  between the WebVision dataset and the ILSVRC 2012 datasets. Nevertheless, the model trained on the WebVision dataset exhibits good generalization ability when the feature representation is transferred to other tasks. It achieves comparable or even better results on the image classification tasks using the Caltech-256 and PASCAL VOC 2007 datasets.
\item \emph{Is the meta information useful for visual recognition?} Another interesting property of web data is the abundant meta information accompanied with images. Such information often provides {a} certain semantic explanation  {of} web images.
\end{itemize}

Besides the above observations on learning from web data, our WebVision database can also be used for other vision problems. For example, since a dataset bias is observed between the WebVision dataset and the ILSVRC 2012 dataset, it can be used as a benchmark setting for the visual domain adaptation task, which to our best knowledge makes it the largest dataset for this task to date. Moreover, the meta-information combined with the training images can also be used for the multi-modality learning tasks. We have released this dataset to the public to advance the research in learning from web data and other related fields~\footnote{\url{http://www.vision.ee.ethz.ch/webvision/index.html}}.

\section{Related Work}
The Internet has been a popular data source for creating various datasets for computer vision research. Many computer vision datasets were constructed by harvesting images from the Internet and filtering with human annotations, including ImageNet~\cite{ImageNet}, PASCAL VOC~\cite{Pascal}, Caltech-256~\cite{Caltech256}, 80M tiny images~\cite{80MImages}, SUN~\cite{SUN}, Places2~\cite{Places2}, MS COCO~\cite{COCO} \etc Moreover, a few video datasets~\cite{ActivityNet,soomro2012ucf101,kuehne2011hmdb} were also built by crawling videos from the Internet. All these datasets rely on humans for further annotation, while the study of this work focuses on directly learning knowledge from noisy web data without using instance-level human annotations.

There is a continued interest in the community on learning visual recognition models directly from web images. Researchers focused on different issues in learning from web data, and have proposed various approaches. Fergus~\etal~\cite{Fergus05} exploited images from the Google search engine for image categorization based on an improved pLSA method. Vijayanarasimhan and Grauman~\cite{Vijayanarasimhan08} proposed a multi-instance learning (MIL) method to explicitly handle the label noise when using web images as training data. Bergamo and Torresani~\cite{bergamo2010exploiting} studied the domain adaptation issue between web images and existing visual recognition benchmark datasets. Schroff~\etal~\cite{Schroff11} proposed an approach to automatically harvest images from the Internet for learning visual recognition classifiers. Li~\etal~\cite{Li14} proposed to exploit the meta information associated with web images to improve the visual recognition performance. However, those works merely studied the problem in small scale, which might not always generalize to large scale problems.  In this work, we conduct extensive experiments with our newly proposed WebVision database, and re-exterminate those issues in a large scale scenario.

\begin{figure*}[t]
\centering
\includegraphics[width=0.9\textwidth, height=180pt]{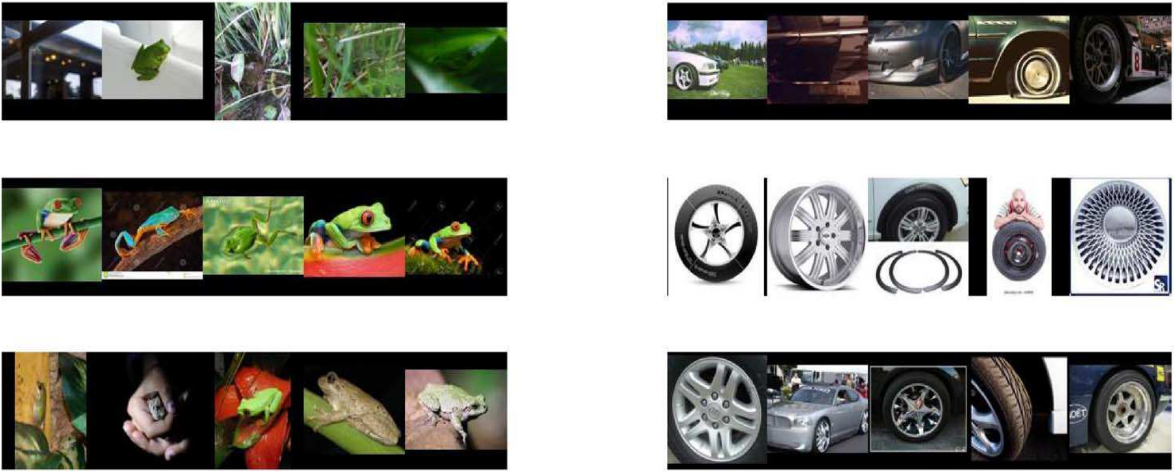}
\caption{Examples of images from Flickr (top), Google (middle), and ImageNet (bottom). Left: ``tree frog''; right: ``car wheel''}
\label{fig:example}
\end{figure*}

A few recent works~\cite{Chen15,Krause16,Joulin16} have also been proposed to utilize web images as training data for learning deep convolutional neural networks (CNNs)~\cite{lecun1998gradient,krizhevsky2012imagenet}. Chen and Gupta~\cite{Chen15} crawled more than $3,000$ concepts of web images, and trained a CNN which achieves comparable results with ImageNet models for object detection and localization. Krause \etal~\cite{Krause16} showed that models learnt from web data outperformed those learnt from human-annotated datasets for find-grained classification tasks. Joulin~\etal~\cite{Joulin16} proposed a word prediction model learnt from web images, and showed good generalization ability for the object detection and scene classification tasks. While those works have shown the power of web data combined with deep learning techniques, their tasks and methodologies differ from each other, making it hard to identify key issues and effective ways when utilizing web data. Moreover, their results were often obtained using much  more  images~\cite{Krause16} or categories~\cite{Chen15}, making it difficult to understand the capacity of noisy web images for learning visual recognition models when compared with the human-annotated datasets. The recent large scale database Google Open images \cite{OpenImages} was announced, but the labels are annotated by a nonpublic machine model and the semantic concepts were not aligned with existing human-annotated datasets. In contrast, the newly proposed WebVision database was built with the same $1,000$ semantic concepts as the popular benchmark human-annotated large scale ILSVRC 2012 dataset. Extensive experiments are conducted in this work to analyze the issues when learning from web data.

From a broader perspective, our work is also related to the recent works on learning visual representation with less human supervision, including unlabeled images/videos~\cite{Vincent10, Radford15,Owens16}, self-supervised and semi-supervised approaches~\cite{Doersch15, Wang15, Agrawal15, Noroozi16}. Learning from web data differs from purely supervised or unsupervised learning because images and videos on the web are naturally accompanied with abundant metadata that can provide weak supervision through search engines without the tedium or expense of crowdsourced manual labels.

\section{WebVision Dataset}
To study learning from web data, we build a large scale web image database called WebVision by crawling web images from the Internet. This new database is then used to investigate the potential of the web data for learning representations in this work. Next, we will describe the details on the construction of the WebVison dataset, and then provide an analysis on it.

\subsection{Dataset Construction}
\textbf{Semantic Concepts:}
The first issue for building a new database is, what semantic concepts of web images shall we collect from the Internet to learn a generic representation? An example of labeled dataset is the ILSVRC 2012 dataset~\cite{ImageNet}, which consists of $1,000$ semantic concepts.  The representation learnt from those $1,000$ concepts of images exhibits good generalization ability, and it has been a common way to fine-tune CNN models learnt from ILSVRC 2012 dataset for various computer vision tasks, such as image classification~\cite{Oquab14}, object detection~\cite{GirshickDDM16}, object segmentation~\cite{ShelhamerLD17} and action recognition~\cite{SimonyanZ14}.  We construct our dataset by collecting web images from the same $1,000$ semantic concepts. Moreover, using the same $1,000$ semantic concepts as the ILSVRC 2012 dataset, it allows us to better understand the potential of the web data for learning representations by directly comparing with ones learnt from the ILSVRC 2012 dataset.

\begin{figure*}[t]
\centering
\includegraphics[width=1.0\textwidth]{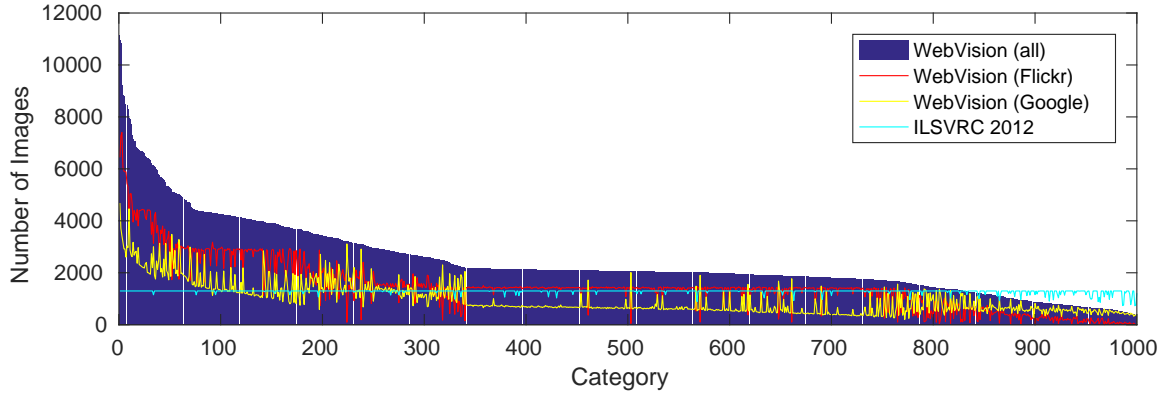}
\caption{Number of images per category of the WebVison dataset.}
\label{fig:hist}
\end{figure*}

\textbf{Web Sources:} We consider two popular sources, the Google Image Search website\footnote{\url{http://images.google.com/}}, and the Flickr website\footnote{\url{http://www.flickr.com/}}. It has been shown in the literature that the images crawled from Google Image Search are effective for image categorization and representation learning~\cite{80MImages,Caltech256,Pascal,Places2,Chen15,Krause16}.

\textbf{Data Collection:}  We individually query images on those two websites. The queries are generated by using the synsets defined in the ILSVRC 2012 dataset.  For the synsets containing multiple items, we treat each item as a query, and crawl images individually for each item in the synset of each category. Items that appear in multiple synsets are revised to avoid conflicts. For example, the synsets of ``n02012849'' and ``n03126707'' are the same, \ie, ``crane''. To eliminate the conflict, we augment those two synsets as ``crane bird'', and ``crane truck, crane tower'', respectively. A complete list of the queries for both website are shown in the Supplementary. In total, we obtain $1,631$ queries from the synsets of $1,000$ semantic categories.

For the Flickr website, we use its text based image search portal, and crawl up to $2,000$ images for each query. We remove images where the short side is less than $500$ pixels, and finally obtain $1.6$M images.

For the Google Image Search website, we crawl as many images as possible for each query, which usually results in $600$--$1,000$ images for each query. After removing the invalid links, we obtained in total $1.1$M images.

For each crawled image, its class label is decided by the synset that its corresponding query belongs to. For example, for the images crawled by using ``crane bird'', its synset ID is ``n02012849'', which has label $135$ using the ILSVRC label set. Since the image search results can be noisy, the training images may contain significant outliers, which is one of the important research issues when utilizing web data (see quantitative results in Section \ref{sec:dataset_anlaysis} and \ref{sec:noise_study}).

\begin{figure}[t]
\centering
\subfigure[Flickr image]{
\includegraphics[width=0.22\textwidth,, height=80pt]{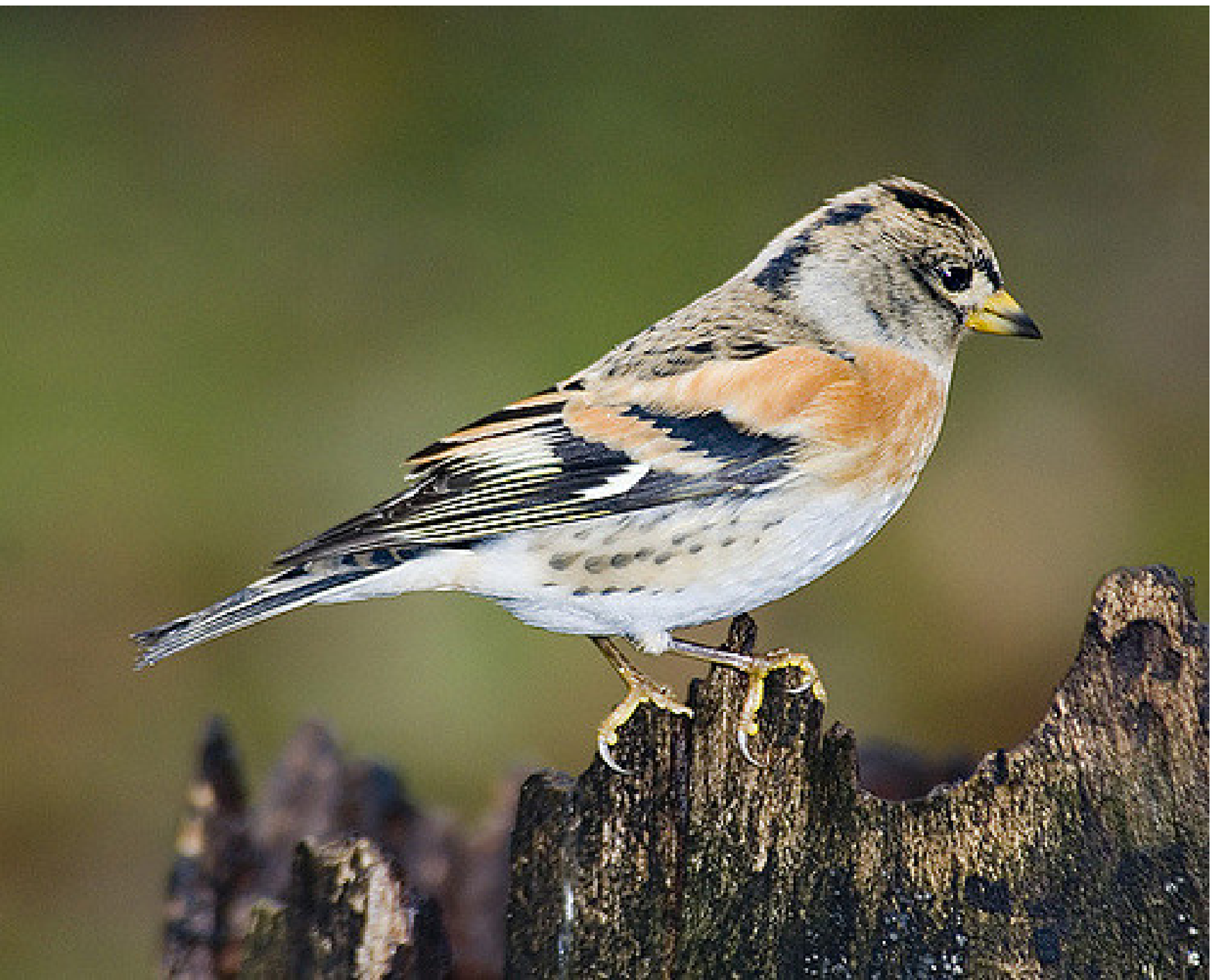}
}
\subfigure[Google image]{
\includegraphics[width=0.22\textwidth, height=80pt]{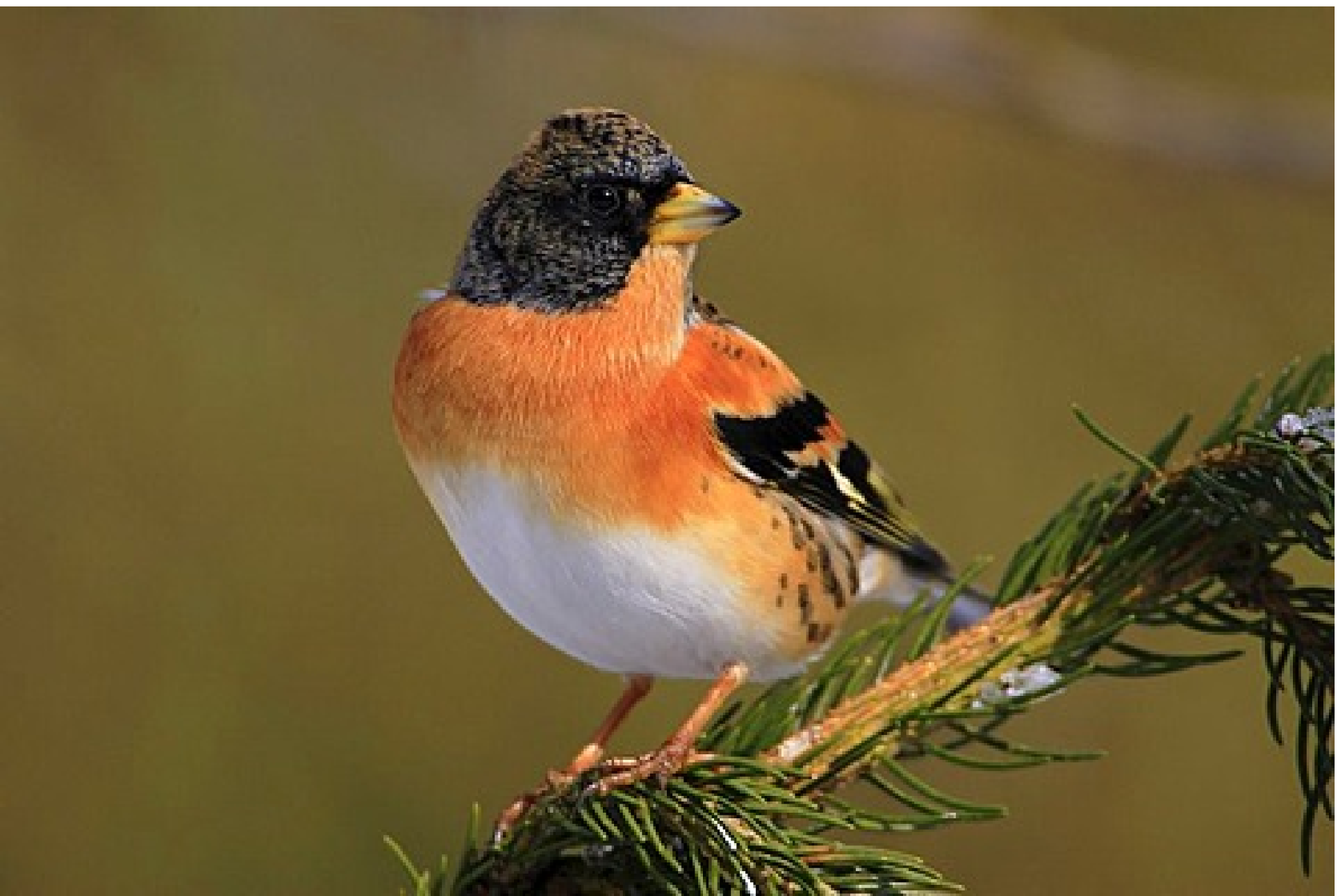}
}
\caption{Examples of image meta information from Flickr and Google. The meta-information associated with these two images is: (a) title: \emph{``Brambling''}; description: \emph{``Brambling - Fringilla montifringilla Russia, Moscow region, Saltykovka, 10/13/2007''}; tags: \emph{"Brambling", "Fringilla montifringilla"}; (b) title: \emph{``High Quality Stock Photos of brambling''}; description:\emph{``Brambling, male, North Rhine-Westphalia, Germany / (Fringilla montifringilla) /''}. }
\label{fig:meta}
\end{figure}

\textbf{Meta Information:} One advantage of web images is the abundant textual information, which usually contains valuable semantic information about the images, and has been shown to be quite useful for image categorization in the literature\cite{Schroff11,Li14,Joulin16}. For each Flickr image, we download its accomplished textual information, including \textit{title}, \textit{description}, \textit{tags}, \etc Geographical information and camera information is also included if it is available. For Google images, the \textit{title} and \textit{description} along with each image are crawled. An example of the meta information associated with images from both sources crawled using the query ``\emph{brambling}'' are shown in Figure~\ref{fig:meta}.

\textbf{Validation and Test Sets:} To facilitate algorithmic development, we also split a subset from the crawled images, and annotate a validation set and a test set. We randomly split out $200,000$ images ($200$ images per category), and put them along with their noisy labels on the Amazon Mechanical Turk (AMT) platform \footnote{\url{http://www.mturk.com/}}. The users are asked to verify if the label provided with each image is correct or not. Each image is annotated by three users, and is considered as an inlier image if more than two users agree. For concepts with less than $100$ inlier images, we continue to split a number of images from the crawled data, and send to AMT for annotation. Finally, we obtain in total $100,000$ human-annotated images, where each of the $1,000$ categories contains $100$ images. We then equally split it into two sets, a validation set and a test set, each containing $50,000$ images, \ie, $50$ images per category.

% \textbf{Near-Duplicate Image Removal:}
The remaining images are used as the training set. To ensure that there is no overlap between the training set and validation or testing set, we perform near-duplicate image detection and remove near duplicate images from the training set~\cite{Places2}.
Specifically, we first resize each image into size of $128 \times 128$ and employ PCA to reduce its dimension to 500. Then, based on the PCA feature, we compute the Euclidean distance of each image pair from training set and validation (or testing) set. Finally, according to the computed distance, we remove the top 5,000 images from the training set, that are most close to the validation data or test data.
Finally, the training set of WebVision database contains in total 2,439,574 images, in which 1,459,125 images are from Flickr and 980,449 images are from Google Image Search.

\subsection{Dataset Analysis}
\label{sec:dataset_anlaysis}
\textbf{Category Distribution: } We plot the number of images per category for our WebVision database as well that for the ILSVRC 2012 dataset in Figure~\ref{fig:hist}. The number of images per category in the ILSVRC 2012 dataset is restricted no more than 1,300. For our WebVision database, and the number of images per category varies from $300$ to more than $10,000$. the number of images per category depends on both the number of queries generated from the synset for each category, and also the availability of images on Flickr and Google. Usually a category with many queries contains more images.

\begin{figure}[t]
\centering
\includegraphics[width=0.44\textwidth]{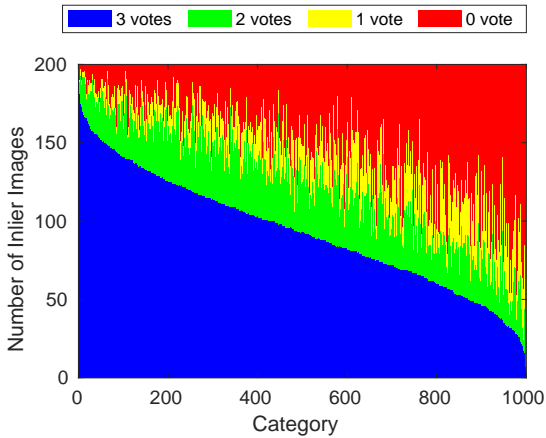}
\caption{Number of inlier images among 200 images per category of the WebVison dataset, sorted by number of ``3 votes" images in descend order. }
\label{fig:truth_num}
\end{figure}

\textbf{Domain Difference: } We give some examples of Flickr and Google images in our WebVision database in Figure~\ref{fig:example}. The images of corresponding categories from the ILSVRC 2012 dataset are also included for comparison. Generally, the Google images are usually with a clean background, and the objects/targets in the image are captured with a clear shot. In contrast, the images from Flickr are usually captured with various backgrounds in the wild, and the objects/targets are sometimes with small sizes. As a comparison, the ILSVRC 2012 dataset is filtered with human annotation, so the objects/targets are usually clearly visible with diverse backgrounds. A quantitative analysis on the domain difference between WebVision and ILSVRC 2012 datasets are given in Section~\ref{sec:dataset_anlaysis}.

\textbf{Noisy Labels:} To investigate how noisy the labels of web images are, we take the annotation results from the first round (200K images) as an example, and plot the user votes in Figure \ref{fig:truth_num}. Each vote indicates that a user agrees the provided label is correct, and images with more than 2 out of 3 votes are considered as true inlier images.

From the figure, we observe that the crawled web images contain a considerable amount of outliers. About 20\% of images are considered as true noisy images (\ie, ``0 vote''), and the  inlier images (\ie, ``3 votes'' and ``2 votes'') take only 66\% of the total images.
Moreover, the number of inlier images varies a lot in different categories.
The  cleanest category is ``867 -- Tractor'' which contains $199$ inlier images among $200$ split images. The worst one is ``627 -lighter, light, igniter, ignitor'', which has only $24$ inlier images.

\section{Experimental Studies}
Our WebVision database contains noisy but rich web data for learning representations, which could be adapted for many high-level vision tasks. In this section, we present a simple baseline for directly learning from web data, and investigate the capacity of the WebVision dataset for different visual recognition tasks by comparing with the human-annotated ILSVRC 2012 dataset.

\subsection{Baseline Model and Dataset Bias}
\label{sec:domain_bias}
{\bf Baseline:} We first propose a simple baseline method to learn visual representation on the training set of our WebVision dataset. Specifically, we treat the query concepts as the semantic label of each image and train an AlexNet model~\cite{krizhevsky2012imagenet}. Following the standard training pipeline, we first resize each image to make shorter size as 256. Then, a patch of size $227 \times 227$ is randomly cropped from each image, and this crop (or its horizontal flipping) is fed into the network for training. We use the mini-batch stochastic gradient algorithm to learn the network parameters, where the batch size is set to 256 and the momentum set to 0.9. The learning rate is initialized as 0.1 and decreased by a factor $\frac{1}{10}$ every $200,000$ iterations, and the whole training procedure stops at $900,000$ iterations. After the model training, we verify the performance of the learned model on the validation data of WebVision and ILSVRC 2012 with a single crop test.

{\bf Results:} We first report the performance of our learned models on the validation data of the WebVision dataset and the results are reported in Table~\ref{tbl:classification_baseline}. We see that our learned AlexNet model obtains the performance of top1 accuracy of 57.03\% and top5 accuracy of 77.90\% on the validation set. As the semantic concepts of our WebVision are the same as in ILSVRC 2012, we can perform cross-dataset testing on the models learned from the WebVision and ILSVRC 2012 dataset. The performance comparison is reported in Table~\ref{tbl:classification_baseline}. We notice that there is a performance drop when we conduct cross-dataset testing, for example, the top5 accuracy of our WebVision model decreases to 70.36\% from 77.90\% and for the ILSVRC 2012 model it decreases from 79.77\% to 74.64\%. These results indicate that there is a domain difference between our WebVision dataset and the ILSVRC 2012  dataset~\cite{torralba2011unbiased,kulis2011you,gopalan2011domain}. Meanwhile, we also notice that the performance drop of our WebVision model is larger than of the ILSVRC 2012 model, which could be explained by the fact that our WebVision dataset is much noisier than the ILSVRC 2012 dataset.

\begin{table}
\caption{Image classification results on the WebVision validation set and the ILSVRC 2012 validation set. The number inside {\resp, outside} the parentheses denotes the Top-1 (\resp, Top-5) classification accuracy (\%) using center crops. The best result on each validation set is denoted in bold.}
\vspace{1mm}
\centering
\begin{tabular}{|l|c|c|}
\hline
&  ILSVRC 2012 Val & WebVision Val\\
\hline
ILSVRC 2012 & \textbf{79.77 (56.79) } &  74.64 (52.58) \\
WebVision & 70.36 (47.55) &  \textbf{77.90 (57.03)} \\
\hline
\end{tabular}
\label{tbl:classification_baseline}
\end{table}

\subsection{Quantity \vs Quality}
\label{sec:noise_study}
\begin{figure}[t]
\centering
\includegraphics[width=0.40\textwidth]{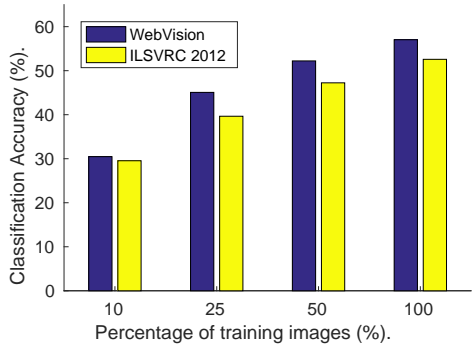}
\caption{Classification accuracy (\%) on WebVision validation set when using different percentages of images in the WebVision dataset and the ILSVRC 2012 dataset.}
\label{fig:acc_ratio}
\end{figure}

\textbf{Image Quantity:} Compared with the ILSVRC 2012 dataset, one obvious advantage of the WebVision dataset is that it contains many more images, around twice the number of images in the ILSVRC 2012 dataset ($2.44$M \vs $1.28$M). To investigate the influence of the number of training images on the recognition performance, we conduct experiments by respectively sampling 10\%, 20\%, 50\% and 100\% of images per category in the WebVision and ILSVRC 2012 datasets. We train AlexNet models, and report their classification accuracy on the WebVision validation set in Figure~\ref{fig:acc_ratio}.

From Figure~\ref{fig:acc_ratio}, we observe that the recognition performance drops for both datasets when reducing the number of training images. We also observe that the models trained from WebVision outperform those from ILSVRC 2012 in all four cases. This is not surprising, considering that the WebVision training images are almost twice as many as those of ILSVRC 2012, and the possible influence from dataset bias. However, when we compare the recognition performance of models using similar number of images, it is interesting to observe that the accuracy using 50\% of WebVision training images is still similar to that using 100\% of ILSVRC 2012 training images ($52.22\%$ \vs $52.58\%$), but the accuracy using 10\% of WebVision training images has a significant gap to that using 25\% of ILSVRC 2012 training images ($30.47\%$ \vs $39.65\%$). We conjecture that this is because the labels of WebVision contain significant noise, which may corrupt the model when the number of training images is limited.

\begin{figure}[t]
\centering
\includegraphics[width=0.40\textwidth]{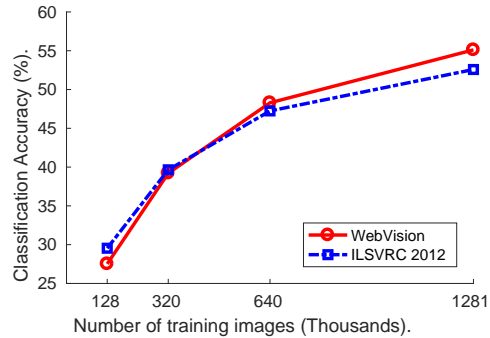}
\caption{Top-1 classification accuracy (\%) on WebVision validation set when using different number of images in the WebVision dataset and the ILSVRC 2012 dataset. The numbers of images correspond to 10\%, 25\%, 50\%, and 100\% of the total number of images in the ILSVRC 2012 dataset, respectively. The WebVision images are sampled such that the number of images per category is as the same as the ILSVRC 2012 dataset.}
\label{fig:acc_num}
\end{figure}

\textbf{Quantity \vs Quality:} To further investigate the impact of label noise in the WebVision dataset, we conduct experiments by respectively sampling $128$K, $320$K, $640$K, and $1.28$M images (corresponding to 10\%, 25\%, 50\%, and 100\% of the total number of ILSVRC 2012 images) from WebVision and ILSVRC 2012. For categories in WebVision with insufficient images, we duplicate the images such that the number of training images per category are ensured the same between two datasets. We train AlexNet models and report in Figure~\ref{fig:acc_num} their classification accuracies on the WebVision validation set.

From Figure~\ref{fig:acc_num}, we observe that the WebVision model achieves better classification accuracy than ILSVRC 2012 dataset when using $1.28$M images (55.11\% \vs 52.58\%). Despite the possible benefits for the WebVision model gained from the dataset bias issue, this result reveals that the harm of label noise issue is neutralized by the large number of training sample. This can be further verified by the results when reducing the number of training images. In this case, the advantage of the WebVision model over the  ILSVRC 2012 model is vanishing, and the recognition accuracy is worse than of the ILSVRC 2012 model when using $128$K training images. This also implies that the visual recognition performance could possibly be further boosted by mining larger number of images from the Internet, which can be done at nearly no cost.

\subsection{Transfer Learning}
To investigate the generalization ability of the AlexNet model learnt from the WebVision dataset, we compare it with the same model learnt from the ILSRVC dataset by transferring the learnt model for image classification and object detection tasks. In the image classification task, we directly evaluate the learnt feature representation on the benchmark Caltech-256 dataset~\cite{Caltech256} and the PASCAL VOC 2007 dataset~\cite{Pascal}. In the object detection task, we perform object detection on the PASCAL VOC 2007 dataset by fine-tuning the pre-learnt models.
\begin{table}
\caption{Image classification results on the Caltech-256 and PASCAL VOC 2007 datasets. The classification accuracy (\%) is reported on the Caltech-256 dataset, and the mean of average precisions (mAP in \%) over 20 categories is reported on PASCAL VOC 2007 dataset.}
\label{tbl:classification}
\vspace{1mm}
\centering
\begin{tabular}{|l|c|c|}
\hline
& Caltech-256 & PASCAL VOC 2007 \\
\hline
ILSVRC 2012 &  70.44 & 75.65\\
WebVision &  70.43 & 77.78\\
\hline
Combined &  \textbf{73.61} & \textbf{78.46} \\
\hline
\end{tabular}
\end{table}
\subsubsection{Image Classification}
\textbf{Experimental Setup:} We use the Caltech-256 and PASCAL-VOC 2007 datasets as the benchmarks to evaluate the feature representation learnt using our WebVision dataset, and compare to the feature representation learnt with the same AlexNet architecture using the ILSVRC dataset.

For the Caltech-256 dataset, 30 images per category are used as the training set and the rest as the testing set. For the PASCAL VOC 2007 dataset, we combine the official train and validation splits as the training set, and use the test split as the test set.

For all CNN models, we use the 4096-d output from the "fc7" layer as the feature representation. Each image is resized such that its short side is 256. Then, ten 227x227 patches are obtained by cropping and flipping five patches at center and four corners. Those ten feature vectors are then averaged as the final feature vector for each image. $\ell_2$ normalization is performed on the final feature vectors.

For the Caltech-256 dataset,  we run ten rounds of random sampling and train a multi-class SVM classifier using the training set at each round. The mean classification accuracy on the test set over ten rounds is reported. For the PASCAL VOC 2007 dataset, since images may have multiple labels, we train a one-versus-all binary SVM classifier for each category, and report the mean of average precision (mAP). The trade-off parameter C for SVM is set as 1 in all experiments, which generally gives the best test results for all models.

\textbf{Results:}
The results of all methods are reported in Table~\ref{tbl:classification}. The result using the feature representation learnt from our WebVision dataset is on par with that from the ILSVRC 2012 dataset on the Caltech-256 dataset (70.43 \vs 70.44), and is better on the PASCAL VOC 2007 dataset (77.78 \vs 75.65). This indicates that the representation learnt directly from web images can generalize to the third dataset as well as or slightly better than the human-annotated ILSVRC 2012 dataset, which clearly demonstrates the good generalization ability of the model learnt using web images for other computer vision tasks.

We fuse the models from our WebVision dataet and the ILSVRC dataset by concatenating two feature representations. The classification accuracy are further boosted  $(70.44, 70.43) \rightarrow 73.61$ on Caltech-256 and $(75.65, 77.78) \rightarrow 78.46$ on PASCAL VOC 2007. This indicates that the WebVision model is complementary to the ILSVRC 2012 model to some extent, and the vision tasks can be benefited by further leveraging abundant web images from the Internet.

\subsubsection{Object Detection}
\begin{table*}[t]
\caption{Object detection results on the PASCAL VOC 2007 dataset. We fine tune the models on the train+val data of PASCAL VOC 2007 by using the Faster-RCNN detection framework, and perform detection on the test set of PASCAL VOC 2007. The average precision (\%) for each category as well as the mean average precision over 20 categories are reported.}
\vspace{1mm}
\centering
\resizebox{\textwidth}{!}{
\setlength{\tabcolsep}{2pt}
\begin{tabular}{l|cccccccccccccccccccc|c}
\hline
{Model} & areo & bike & bird & boat & bottle & bus & car & cat & chair & cow & table & dog & horse & mbike & person & plant & sheep & sofa & train & tv & {mAP} \\ \hline
{ILSVRC 2012} & 64.0 & 67.3 & 53.5 & 39.6 & 32.4 & 66.6 & 71.6 & 66.4 & 36.6 & 61.1& 58.4 & 62.2 & 75.1 & 70.4 & 65.0 & 33.1 & 56.5 & 48.0 & 69.3 & 64.4 & \textbf{58.1} \\
{WebVision} & 64.6 & 70.6 & 50.8 & 41.8 & 28.6 & 66.8 & 71.4 & 69.4 & 34.6 & 63.2 & 61.8 & 62.1 & 74.4 & 69.7 & 65.1 & 32.8 & 53.2 & 52.2 & 70.8 & 59.5 & \textbf{58.2} \\ \hline
\end{tabular}
}
\label{tbl:pascal07-det}
\end{table*}
We further investigate the generalization ability WebVision model when being fine-tuned to the new vision tasks.  We take the object detection task as an example by using the state-of-the-art Faster-RCNN framework~\cite{ren2015faster} on the PASCAL VOC 2007 detection dataset. Following the standard Faster-RCNN framework~\cite{ren2015faster}, we use the PASCAL VOC 2007 train/val images to fine-tune the detection networks by using the AlexNet models trained from the WebVision and ILSVRC 2012 datasets, respectively. The experiments are conducted using the  existing public faster-rcnn toolbox \footnote{Codes are available at https://github.com/rbgirshick/py-faster-rcnn}. The object detection performance is measured on the PASCAL VOC 2007 test images. The results are evaluated by average precision (AP) and mean AP is reported to measure the overall detection performance.

The object results are summarized in Table \ref{tbl:pascal07-det}. We observe that the WebVision model achieves comparable results compared with the ILSVRC 2012 model, which demonstrates its good generalization ability when being transferred to a new
vision task in a fine-tuning fashion.

\section{Conclusions and Future Work}
In this work, we have studied the problem of learning visual recognition models from large scale noisy web data. A new large scale web images database called \emph{WebVision} was built, which contains more than $2.4$ million web images crawled from Flickr and Google Image Search by using queries generated from the same $1,000$ synsets as the benchmark ILSVRC 2012 dataset. Meta information along with those web images (\eg, title, description, tags, \etc) were also crawled. We further annotated a validation set and test set to facilitate algorithmic development.

Based on this dataset, we conducted extensive experiments which gave a few interesting observations:
\begin{itemize}
\setlength{\itemsep}{0pt}
\setlength{\parskip}{0pt}
\setlength{\parsep}{0pt}
\item Considerable label noise exists in the web images, but the WebVision database is still able to train a robust deep CNN model for visual recognition. The large number of training images overcome the noisy label issue in the web data.
\item The visual knowledge learnt from the WebVision can be  transferred to new datasets and tasks as effectively as from the  ILSVRC 2012 dataset. Comparable or even better results were achieved for image classification on Caltech-256 and PASCAL VOC 2007, and for object detection on PASCAL VOC 2007.
\item A domain difference was observed between the WebVision database and the ILSVRC 2012 dataset, which is verified with both qualitative and quantitative results.
\end{itemize}

Based on the newly proposed dataset, an incomplete {list of} potentially interesting research issues in related fields could be: 1) Although the label noise issue is alleviated by the large number of training images to some extent, it is interesting to develop new deep learning approaches to cope with the noisy labels in web images to further boost the visual recognition performance; 2) Considering the dataset bias between the WebVision and ILSVRC 2012 dataset, it would also be interesting to utilize our new dataset to study the visual domain adaptation issue in a large scale and loosely labeled scenario; 3) It is also desirable to study the effectiveness of other types of meta-information such as ``title'', ``tag'', ``description'', \etc
New algorithms for effectively incorporating the meta-information when learning deep models are also in high demand.

\section*{Acknowledgement}
We sincerely thank Prof. Rahul Sukthankar for his valuable discussion in building this dataset. This work is suppored by the Computer Vision Laboratory at ETH Zurich, and the Google Research Europe. The authors gratefully thank NVIDIA Corporation for donating the GPUs used in this project.


{\small
\begin{thebibliography}{10}\itemsep=-1pt
\bibitem{Agrawal15}
P.~Agrawal, J.~Carreira, and J.~Malik.
\newblock Learning to see by moving.
\newblock In {\em ICCV}, 2015.

\bibitem{bergamo2010exploiting}
A.~Bergamo and L.~Torresani.
\newblock Exploiting weakly-labeled web images to improve object
  classification: a domain adaptation approach.
\newblock In {\em Advances in Neural Information Processing Systems}, pages
  181--189, 2010.

\bibitem{Chen15}
X.~Chen and A.~Gupta.
\newblock Webly supervised learning of convolutional networks.
\newblock In {\em ICCV}, 2015.

\bibitem{ImageNet}
J.~Deng, W.~Dong, R.~Socher, and L.~Fei-Fei.
\newblock Imagenet: a large-scale hierarchical image database.
\newblock In {\em CVPR}, 2008.

\bibitem{Doersch15}
C.~Doersch, A.~Gupta, and A.~A. Efros.
\newblock Unsupervised visual representation learning by context prediction.
\newblock In {\em ICCV}, 2015.

\bibitem{DongLHT16}
C.~Dong, C.~C. Loy, K.~He, and X.~Tang.
\newblock Image super-resolution using deep convolutional networks.
\newblock {\em {IEEE} Trans. Pattern Anal. Mach. Intell.}, 38(2):295--307,
  2016.

\bibitem{DosovitskiyFIHH15}
A.~Dosovitskiy, P.~Fischer, E.~Ilg, P.~H{\"{a}}usser, C.~Hazirbas, V.~Golkov,
  P.~van~der Smagt, D.~Cremers, and T.~Brox.
\newblock Flownet: Learning optical flow with convolutional networks.
\newblock In {\em ICCV}, pages 2758--2766, 2015.

\bibitem{Pascal}
M.~Everingham, L.~{Van Gool}, C.~Williams, J.~Winn, and A.~Zisserman.
\newblock The {PASCAL} visual object classes ({VOC}) challenge.
\newblock {\em IJCV}, 88(2):303--338, 2010.

\bibitem{ActivityNet}
B.~G. Fabian Caba~Heilbron, Victor~Escorcia and J.~C. Niebles.
\newblock Activitynet: A large-scale video benchmark for human activity
  understanding.
\newblock In {\em Proceedings of the IEEE Conference on Computer Vision and
  Pattern Recognition}, pages 961--970, 2015.

\bibitem{Fergus05}
R.~Fergus, L.~Fei-Fei, P.~Perona, and A.~Zisserman.
\newblock Learning object categories from google’s image search.
\newblock In {\em ICCV}, 2005.

\bibitem{GirshickDDM16}
R.~B. Girshick, J.~Donahue, T.~Darrell, and J.~Malik.
\newblock Region-based convolutional networks for accurate object detection and
  segmentation.
\newblock {\em {IEEE} Trans. Pattern Anal. Mach. Intell.}, 38(1):142--158,
  2016.

\bibitem{gopalan2011domain}
R.~Gopalan, R.~Li, and R.~Chellappa.
\newblock Domain adaptation for object recognition: An unsupervised approach.
\newblock In {\em Computer Vision (ICCV), 2011 IEEE International Conference
  on}, pages 999--1006. IEEE, 2011.

\bibitem{Caltech256}
G.~Griffin, A.~Holub, and P.~Perona.
\newblock Caltech-256 object category dataset.
\newblock Technical report, California Institute of Technology, 2007.

\bibitem{HeZRS16}
K.~He, X.~Zhang, S.~Ren, and J.~Sun.
\newblock Deep residual learning for image recognition.
\newblock In {\em CVPR}, pages 770--778, 2016.

\bibitem{Joulin16}
A.~Joulin, L.~van~der Maaten, A.~Jabri, and N.~Vasilache.
\newblock Learning visual features from large weakly supervised data.
\newblock In {\em ECCV}, 2016.

\bibitem{OpenImages}
I.~Krasin, T.~Duerig, N.~Alldrin, A.~Veit, S.~Abu-El-Haija, S.~Belongie,
  D.~Cai, Z.~Feng, V.~Ferrari, V.~Gomes, A.~Gupta, D.~Narayanan, C.~Sun,
  G.~Chechik, and K.~Murphy.
\newblock Openimages: A public dataset for large-scale multi-label and
  multi-class image classification.
\newblock {\em Dataset available from https://github.com/openimages}, 2016.

\bibitem{Krause16}
J.~Krause, B.~Sapp, A.~Howard, H.~Zhou, A.~Toshev, T.~Duerig, J.~Philbin, and
  L.~Fei-Fei.
\newblock The unreasonable effectiveness of noisy data for fine-grained
  recognition.
\newblock In {\em ECCV}, 2016.

\bibitem{krizhevsky2012imagenet}
A.~Krizhevsky, I.~Sutskever, and G.~E. Hinton.
\newblock Imagenet classification with deep convolutional neural networks.
\newblock In {\em Advances in neural information processing systems}, pages
  1097--1105, 2012.

\bibitem{kuehne2011hmdb}
H.~Kuehne, H.~Jhuang, E.~Garrote, T.~Poggio, and T.~Serre.
\newblock Hmdb: a large video database for human motion recognition.
\newblock In {\em Computer Vision (ICCV), 2011 IEEE International Conference
  on}, pages 2556--2563, 2011.

\bibitem{kulis2011you}
B.~Kulis, K.~Saenko, and T.~Darrell.
\newblock What you saw is not what you get: Domain adaptation using asymmetric
  kernel transforms.
\newblock In {\em Computer Vision and Pattern Recognition (CVPR), 2011 IEEE
  Conference on}, pages 1785--1792. IEEE, 2011.

\bibitem{lecun1998gradient}
Y.~LeCun, L.~Bottou, Y.~Bengio, and P.~Haffner.
\newblock Gradient-based learning applied to document recognition.
\newblock {\em Proceedings of the IEEE}, 86(11):2278--2324, 1998.

\bibitem{Li14}
W.~Li, L.~Niu, and D.~Xu.
\newblock Exploiting privileged information from web data for image
  categorization.
\newblock In {\em ECCV}, 2014.

\bibitem{COCO}
T.-Y. Lin, M.~Maire, S.~Belongie, J.~Hays, P.~Perona, D.~Ramanan, P.~Dollar,
  and C.~L. Zitnick.
\newblock Microsoft coco: Common objects in context.
\newblock In {\em ECCV}, 2014.

\bibitem{long2015fully}
J.~Long, E.~Shelhamer, and T.~Darrell.
\newblock Fully convolutional networks for semantic segmentation.
\newblock In {\em Proceedings of the IEEE Conference on Computer Vision and
  Pattern Recognition}, pages 3431--3440, 2015.

\bibitem{Noroozi16}
M.~Norooz and P.~Favaro.
\newblock Unsupervised learning of visual representations by solving jigsaw
  puzzles.
\newblock In {\em ECCV}, 2016.

\bibitem{Oquab14}
M.~Oquab, L.~Bottou, I.~Laptev, and J.~Sivic.
\newblock Learning and transferring mid-level image representations using
  convolutional neural networks.
\newblock In {\em 1717--1724}, page June, 2014.

\bibitem{Owens16}
A.~Owens, J.~Wu, J.~Mcdermott, A.~Torralba, and W.~Freeman.
\newblock Unsupervised learning of visual representations by solving jigsaw
  puzzles.
\newblock In {\em ECCV}, 2016.

\bibitem{Radford15}
A.~Radford, L.~Metz, and S.~Chintalan.
\newblock Unsupervised representation learning with deep convolutional
  generative adversarial networks.
\newblock In {\em arXiv:/1511.06434}, 2015.

\bibitem{ren2015faster}
S.~Ren, K.~He, R.~Girshick, and J.~Sun.
\newblock Faster r-cnn: Towards real-time object detection with region proposal
  networks.
\newblock In {\em Advances in neural information processing systems}, pages
  91--99, 2015.

\bibitem{Schroff11}
F.~Schroff, A.~Criminisi, and A.~Zisserman.
\newblock Harvesting image databases from the web.
\newblock {\em T-PAMI}, 33(4):756--766, 2011.

\bibitem{ShelhamerLD17}
E.~Shelhamer, J.~Long, and T.~Darrell.
\newblock Fully convolutional networks for semantic segmentation.
\newblock {\em {IEEE} Trans. Pattern Anal. Mach. Intell.}, 39(4):640--651,
  2017.

\bibitem{SimonyanZ14}
K.~Simonyan and A.~Zisserman.
\newblock Two-stream convolutional networks for action recognition in videos.
\newblock In {\em Advances in Neural Information Processing Systems 27: Annual
  Conference on Neural Information Processing Systems 2014, December 8-13 2014,
  Montreal, Quebec, Canada}, pages 568--576, 2014.

\bibitem{soomro2012ucf101}
K.~Soomro, A.~R. Zamir, and M.~Shah.
\newblock UCF101: A dataset of 101 human actions classes from videos in the
  wild.
\newblock {\em arXiv preprint arXiv:1212.0402}, 2012.

\bibitem{SzegedyLJSRAEVR15}
C.~Szegedy, W.~Liu, Y.~Jia, P.~Sermanet, S.~E. Reed, D.~Anguelov, D.~Erhan,
  V.~Vanhoucke, and A.~Rabinovich.
\newblock Going deeper with convolutions.
\newblock In {\em CVPR}, pages 1--9, 2015.

\bibitem{torralba2011unbiased}
A.~Torralba and A.~A. Efros.
\newblock Unbiased look at dataset bias.
\newblock In {\em Computer Vision and Pattern Recognition (CVPR), 2011 IEEE
  Conference on}, pages 1521--1528. IEEE, 2011.

\bibitem{80MImages}
A.~Torralba, R.~Fergus, and W.~T. Freeman.
\newblock 80 million tiny images: a large dataset for non-parametric object and
  scene recognition.
\newblock {\em T-PAMI}, 30(11):1958--1970, 2008.

\bibitem{TranBFTP15}
D.~Tran, L.~D. Bourdev, R.~Fergus, L.~Torresani, and M.~Paluri.
\newblock Learning spatiotemporal features with 3d convolutional networks.
\newblock In {\em ICCV}, pages 4489--4497, 2015.

\bibitem{Vijayanarasimhan08}
S.~Vijayanarasimhan and K.~Grauman.
\newblock Keywords to visual categories: Multiple-instance learning for weakly
  supervised object categorization.
\newblock In {\em CVPR}, 2008.

\bibitem{Vincent10}
P.~Vincent, H.~Larochelle, I.~Lajoie, Y.~Bengio, and P.~Manzagol.
\newblock Stacked denoising autoencoders: Learning useful representations in a
  deep network with a local denoising criterion.
\newblock {\em JMLR}, 11:3371–--3408, 2010.

\bibitem{WangXW0LTG16}
L.~Wang, Y.~Xiong, Z.~Wang, Y.~Qiao, D.~Lin, X.~Tang, and L.~Van Gool.
\newblock Temporal segment networks: Towards good practices for deep action
  recognition.
\newblock In {\em ECCV}, pages 20--36, 2016.

\bibitem{Wang15}
X.~Wang and A.~Gupta.
\newblock Unsupervised learning of visual representations using videos.
\newblock In {\em ICCV}, 2015.

\bibitem{SUN}
J.~Xiao, J.~Hays, K.~Ehinger, A.~Oliva, and A.~Torralba.
\newblock Sun database: Large-scale scene recognition from abbey to zoo.
\newblock In {\em CVPR}, 2010.

\bibitem{Places2}
B.~Zhou, A.~Khosla, {\`{A}}.~Lapedriza, A.~Torralba, and A.~Oliva.
\newblock Places: An image database for deep scene understanding.
\newblock {\em CoRR}, abs/1610.02055, 2016.

\end{thebibliography}
}
\end{document}